\documentclass{bmvc2k}

\usepackage{times}
\usepackage{epsfig}
\usepackage{graphicx}
\usepackage{amsmath}
\usepackage[utf8]{inputenc} 
\usepackage[T1]{fontenc}    
\usepackage{url}            
\usepackage{booktabs}
\usepackage{multirow}
\usepackage{graphicx}
\usepackage{floatrow}
\usepackage{amsfonts}       
\usepackage{algorithm}
\usepackage{algpseudocode}
\usepackage{nicefrac}       
\usepackage{microtype}      
\usepackage{amsmath}
\usepackage{nccmath, mathtools}
\usepackage{enumitem}
\usepackage{floatrow}
\usepackage{wrapfig}
\usepackage{comment}
\usepackage{amssymb}
\usepackage{wrapfig}
\usepackage{xcolor}
\usepackage{lipsum}

\newcommand\blfootnote[1]{%
  \begingroup
  \renewcommand\thefootnote{}\footnote{#1}%
  \addtocounter{footnote}{-1}%
  \endgroup
}

\renewcommand{\thefootnote}{\fnsymbol{footnote}}

\newcommand{\bigtheta}{\boldsymbol{\theta}}

\definecolor{layer1}{RGB}{222,45,38}
\definecolor{layer2}{RGB}{49,130,189}
\definecolor{layer3}{RGB}{49,163,84}
\definecolor{s1}{RGB}{254,153,41}
\definecolor{s2}{RGB}{122,1,119}
\definecolor{s3}{RGB}{221,52,151}
\title{MUSE: Feature Self-Distillation with Mutual Information and Self-Information}

\addauthor{Yu Gong\textsuperscript{*}}{gongyug@sfu.ca}{2}
\addauthor{Ye Yu\textsuperscript{*}}{yu.ye@microsoft.com}{1}
\addauthor{Gaurav Mittal}{gaurav.mittal@microsoft.com}{1}
\addauthor{Greg Mori}{mori@cs.sfu.ca}{2}
\addauthor{Mei Chen}{mei.chen@microsoft.com}{1}

\addinstitution{
 Microsoft
}
\addinstitution{
 Simon Fraser University
}

\runninghead{Y. Gong, Y. Yu, G. Mittal, G. Mori, and M. Chen}{MUSE}

\begin{document}

\maketitle

\begin{abstract}
   We present a novel information-theoretic approach to introduce dependency among features of a deep convolutional neural network (CNN). The core idea of our proposed method, called MUSE, is to combine MUtual information and SElf-information to jointly improve the expressivity of all features extracted from different layers in a CNN.
   We present two variants of the realization of MUSE---Additive Information and Multiplicative Information.  
   Importantly, we argue and empirically demonstrate that MUSE, compared to other feature discrepancy functions, is a more functional proxy to introduce dependency and effectively improve the expressivity of all features in the knowledge distillation framework. MUSE achieves superior performance over a variety of popular architectures and feature discrepancy functions for self-distillation and online distillation, and performs competitively with the state-of-the-art methods for offline distillation.
   MUSE is also demonstrably versatile that enables it to be easily extended to CNN-based models on tasks other than image classification such as object detection.\blfootnote{* Authors with equal contribution. Work was done when Yu Gong was a research intern at Microsoft.} 
\end{abstract}

\setlength{\textfloatsep}{0.15cm}
\vspace{-1em}
\section{Introduction}\label{intro}
\vspace{-0.5em}
There has been extensive research on convolutional neural networks (CNNs) for computer vision tasks.
A variety of deep convolutional network architectures have emerged, whether empirically designed such as VGG~\cite{simonyan2015deep}, ResNet~\cite{resnet}, WideResNet~\cite{Zagoruyko2016WRN}, DenseNet~\cite{huang2017densely}, ShuffleNet~\cite{8578814}, or constructed using Neural Architecture Search such as NASNet~\cite{zoph2018learning} and the baseline network of EfficientNet~\cite{pmlr-v97-tan19a}. 
While the large number of network weights store the experience learned from the training data, the activations (or \textit{features}) of the hidden layers represent the direct response from the network to the data. 
It is still an open problem how to fully utilize these intermediate features to advance the capacity of the models. In this work, we aim to improve the existing neural architectures by fully exploiting the features.

Based on provably effective neural estimator on mutual information~\cite{pmlr-v80-belghazi18a}, recent progress~\cite{Tschannen2020, hjelm2018learning} on unsupervised representation learning treat features as random variables and formulate the \textit{information} of the features to learn a maximally informative representation of the data.
\textit{Mutual Information} (MI) is an information-theoretic measure to quantify the amount of information shared by two random variables.
We posit that \textit{Self-information} (SI), which quantifies the amount of information in one random variable, also plays an essential role. SI can be viewed as MI between two identical random variables, which enables us to estimate it using MI neural estimators.
Instead of learning one single effective data representation, we aim to estimate and combine the MI and SI of multiple intermediate features to further boost the discriminative power of deep CNNs.

\textit{Knowledge distillation} (KD)~\cite{hinton2015distilling} is the practice that was proposed to learn a comparably performant compact neural network from a powerful yet expensive teacher network without additional architecture modification. The core idea is to let the compact student network mimic the ``soft targets'' produced by the teacher network.
It can be further divided into \textit{{offline distillation}} where the teacher network is pretrained and fixed, and \textit{{online distillation}} where the teacher and student networks are jointly trained from scratch.
Unlike offline and online distillation, \textit{self-distillation}  (SD) does not involve multiple networks, but aims to learn by distilling its own knowledge. It is self-contained as it does not require an extra teacher with additional training overhead and removes the need for teacher model selection while focusing solely on the target model. Our work is under this category as we aim to improve the overall performance by the intermediate features in one single network. 
Prior work on SD usually relies on the penalty function proposed to distill knowledge between two networks. By treating the features as random variables, we propose a novel discrepancy function based on {{MUtual information and SElf-information}}, called \textbf{\textit{MUSE}}, to better self-distill the knowledge from different features extracted within one network and improve each feature. 
We validate this approach on various backbone CNN networks on image classification and object detection and show its effectiveness.
We summarize our contributions as follows:

\begin{itemize}[leftmargin=*]
    \item A novel feature discrepancy function MUSE with two realization variants, \textit{Additive Information} and \textit{Multiplicative Information}, to introduce strong dependencies among features within a CNN. We demonstrate its effectiveness in feature distillation \textcolor{black}{and how self-information (SI) interacts with mutual information (MI) to improve distillation.}
    \item Outperforming other state-of-the-art self-distillation (SD) methods when applying SD framework on image classification and object detection, indicating the ability of the proposed method to enhance the feature expressivity.
    \item Establishing the efficacy for model compression, where the compressed models perform competitively or even better than the original architecture while significantly reducing parameters and computation.
    \item Validating the general applicability of MUSE on online distillation and offline distillation.
\end{itemize}

\vspace{-1em}
\section{Related Work}
\vspace{-0.5em}

\paragraph{Mutual Information Estimation.} MI is a widely used information-theoretic measure to quantify the amount of information shared by two random variables, defined as a
Kullback–Leibler (KL) divergence $\mathcal{I}(X; Y) = D_{\text{KL}}(p(x,y)||p(x)p(y))$.
It performs as a measure of true dependency between random variables~\cite{Kinney3354, pmlr-v80-belghazi18a}. However, the exact computation of MI between continuous variables is intractable. Traditional parameteric~\cite{valhu2005} and non-parameteric~\cite{PhysRevE.69.066138,PhysRevE.69.056111} estimation methods are hard to scale up to high dimensionality or large sample size. Mutual Information Neural Estimator (MINE~\cite{pmlr-v80-belghazi18a}) presents a scalable parametric neural estimator based on the dual representation of the KL divergence. 
MI is demonstrated in unsupervised representation learning to learn useful representations~\cite{NEURIPS2019_ddf35421, hjelm2018learning,Tschannen2020}. These methods typically attempt to maximize a pair-wise MI between the input and the representations.
 In practice, MI is often estimated and maximized in a multi-view formulation of the input to allow more modeling flexibility (see~\cite{Tschannen2020}). 
The \textit{views}~\cite{tian2020contrastiveCoding} of the data are chosen from prior knowledge to capture entirely different aspects, e.g., different image channels (luminance, chrominance, and depth)~\cite{tian2020contrastiveCoding} and an autoregressive manner of the patches~\cite{oord2019representation}. We aim to use MI to introduce dependencies between multiple intermediate features to combine information from different levels. Thus, we follow Deep InfoMax (DIM)~\cite{hjelm2018learning} to construct the views of \textit{global} and \textit{local} structures. The global structure captures the summarized information while the local structure models the structural information (e.g. spatial locality). \\
\textbf{Knowledge Distillation.}
Larger CNNs have been shown to achieve higher accuracy as over-parameterization brings more learning capacity to generalize to new data. Hence, {Knowledge Distillation}~\cite{hinton2015distilling} was originally proposed to train a compact student network supervised by the logits from a larger teacher network.
The ``knowledge'' can be captured by the KL divergence between logits~\cite{hinton2015distilling}, or different feature discrepancy functions on the intermediate feature maps, e.g., L2 loss~\cite{Romero15fitnets:hints, Zhang_2019_ICCV}, adversarial loss~\cite{chung2020feature, shen2019meal}, Maximum Mean Discrepancy (MMD)~\cite{NST2017}, \textit{etc}.
Based on the learning scheme, knowledge distillation can be divided into three main categories: 1) {Offline distillation}~\cite{hinton2015distilling, Romero15fitnets:hints, Tian2020Contrastive}, where the pretrained teacher network distills its knowledge to supervise the student training; 2) {Online distillation}~\cite{zhang2017deep, chung2020feature, Guo_2020_CVPR}, where the teacher and student models are both trained from scratch and the knowledge is distilled simultaneously; 3) Self-distillation~\cite{Zhang_2019_ICCV, yang2019snapshot, ddsgd2019, Yun_2020_CVPR}, where one single network is trained to distill its own knowledge into itself. In our work, we argue and empirically demonstrate that the proposed feature discrepancy function based on MI and SI can significantly improve SD. We also show its effectiveness when applying our method to offline and online distillation.

\vspace{-1em}

\begin{figure}[t]
\begin{center}

   \includegraphics[width=0.9\textwidth]{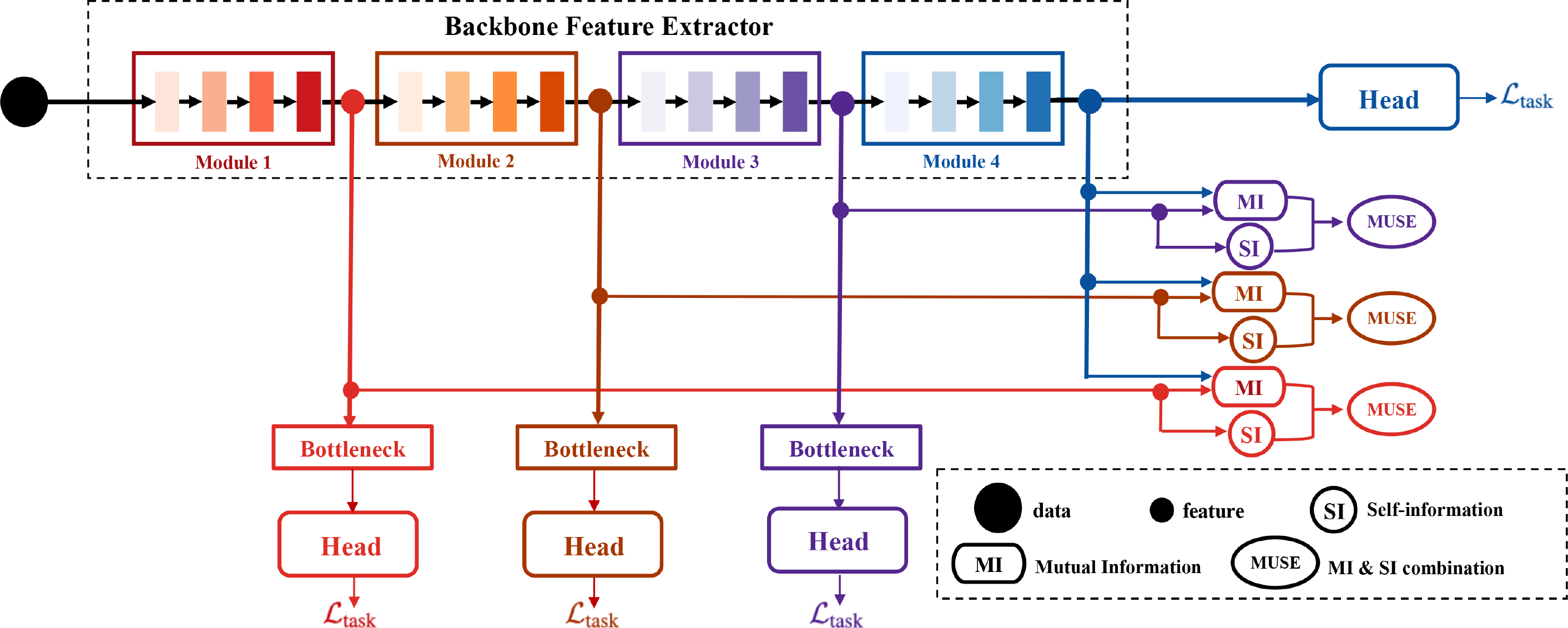}

\end{center}
\vspace{-1em}
\caption{\textbf{Illustration of self-distillation framework with MUSE.} Each color denotes a different module with a task-specific head. MUSE is calculated between  Module 1-3 and Module 4. The entire network is jointly trained with MUSE and a task-specific loss.}
\label{fig::arch}\vspace{-1em}
\end{figure}

\section{Methodology}
\vspace{-0.5em}
We aim to leverage MI and SI to introduce strong feature dependencies and enhance feature expressivity in a CNN, thereby effectively improving the distillation frameworks on different tasks with various backbone CNN networks.  At a high level, our self-distillation (SD) framework relies on multiple (three in Fig.~\ref{fig::arch}) intermediate features and the last feature (e.g., features before fully-connected layers in classification networks). For each intermediate feature, we calculate two components---task-specific objective $\mathcal{L}_\text{task}$ and our proposed MUSE. MUSE is an effective feature discrepancy function to introduce dependencies and enhance expressivity of all features, thereby improving the overall performance and also obtaining compact yet comparably performant subnetworks.

\vspace{-1em}
\subsection{Notation and Preliminaries}
\vspace{-0.5em}
\label{sec:prelis}
\textbf{Mathematical Formulation.}
A CNN
can be parameterized by $\{ \bigtheta_1,\bigtheta_2, \cdots, \bigtheta_T \}$, where $T$ is the length of depth-wise decomposition and each $\bigtheta_i$ contains multiple consecutive hidden layers. 
Let $ X \thicksim p_{\text{data}}(x)$
be a random sample drawn from the empirical data distribution, the \textit{feature} at module $t$ is obtained by a nonlinear transformation $ F_t = E_{\bigtheta_{<t+1}}(X)$.\medskip\\
\textbf{Mutual Information between features.}
For the features $F_i$ and $F_j$ ($i < j$), the MI can be defined as conditional entropy $\mathcal{H}(F_j|F_i)$ subtracted from the self-information $\mathcal{H}(F_j)$, 
\setlength{\belowdisplayskip}{5pt} \setlength{\belowdisplayshortskip}{5pt}
\setlength{\abovedisplayskip}{5pt} \setlength{\abovedisplayshortskip}{5pt}
\begin{align}\label{eq:mi_layers}
    \mathcal{I}(F_{i}; F_{j}) = \mathcal{H}(F_j) - \mathcal{H}(F_j|F_i)
\end{align}

\textbf{Self-Information in features.}
SI (or entropy) quantifies the amount of information of one random variable, and can be defined as MI between identical variables. The SI of $F_i$ is,
\begin{align}\label{eq:self_info}
\mathcal{H}(F_i) = \mathcal{H}(F_i) - \mathcal{H}(F_i|F_i) = \mathcal{I}(F_{i}; F_{i}) 
\end{align}

\vspace{-1.5em}
\subsection{Proposed Method}
\vspace{-0.5em}

\label{sec::MUSE}

\paragraph{Introducing feature dependency by Mutual Information.} 
Given the decomposition of $T$ modules, we have the features $\{ F_1, F_2, \cdots, F_T\}$. As we aim to enhance the performance of CNNs, we choose the last feature $F_T$ as the base feature and introduce dependency from shallow features $\{F_1, F_2, ..., F_{T-1}\}$ to $F_T$. For any module $i<T$, the pair-wise MI is
\begin{align}\label{eq:mi_each_pair}
        \mathcal{I}(F_i; F_T) = 
        \mathcal{H}(F_T) - \mathcal{H}(F_T|F_i) = 
        \mathcal{H}(F_i) - \mathcal{H}(F_i|F_T)
\end{align}
MI quantifies the dependency between two features. Therefore, we can introduce a strong dependency between each pair of features by optimizing the sum of each $\mathcal{I}(F_i; F_T)$. By maximizing the MI between pairs of each shallow feature and the last feature, the information shared between the feature pairs is captured and maximized. In CNNs, lower layers usually learn features of simple patterns, whereas upper layers learn more invariant and global features~\cite{DB15a}. The intermediate features $F_1,...,F_{T-1}$ can gain the global information from the most parameterized and expressive feature $F_T$ within the CNN architecture. Meanwhile, $F_T$ can be aware of more local information that is likely ignored without the introduction of the dependency. To the best of our knowledge, we are the first to use mutual information to introduce the feature dependency for self-distillation in an individual network. \medskip\\
Prior work~\cite{hou2019learning, Zhang_2019_ICCV} typically uses L2 loss to minimize the distance in the feature space. We argue that this practice is not a good proxy to introduce dependency. Minimizing L2 loss is maximizing the likelihood by assuming the data is drawn from a Gaussian distribution. For a pair of features $F_1\sim p(F_1)$ and $F_2\sim q(F_2)$, maximum likelihood estimation (MLE) is known to be equivalent to minimizing the  KL divergence $D_\text{KL}(p||q)$. Therefore, the minimum is obtained when $p=q$. It necessitates two identical distributions. However, in the scenario of self-distillation, we do not expect $p=q$, as they are obtained from different layers within one single network. They depend on each other through non-linear projections, but inherently should not be identical to preserve the semantics of different features learned by the networks. In Section~\ref{sec::ablation}, we empirically demonstrate the effectiveness of MI in comparison to L2 loss.\medskip\\
\textbf{Enhancing feature expressivity by Self-Information.} 
In Eq.~\ref{eq:mi_each_pair}, MI between shallow feature $F_i$ and last feature $F_T$ can be written in two analytically equivalent ways. Since $F_T$ is a non-linear projection from $F_i$, we consider the first form $\mathcal{I}(F_i; F_T) = 
\mathcal{H}(F_T) - \mathcal{H}(F_T|F_i)$ to reflect this casual relationship between $F_i$ and $F_T$. By maximizing $\mathcal{I}(F_i; F_T)$, $\mathcal{H}(F_T)$ drives $F_T$ to spread in the feature space, and the conditional entropy $\mathcal{H}(F_T|F_i)$ enforces $F_T$ to be easily identified given corresponding shallow feature $F_i$. Besides maximizing SI of $F_T$, we also aim to explicitly maximize the SI of shallow features.
To this end, we introduce two realizations of MUSE to combine the MI and SI as follows.\medskip\\
\textbf{\textit{Additive Information.}} \textcolor{black}{The intuition originates from the two forms of MI (Eq.~\ref{eq:mi_each_pair}). Maximizing pure MI is maximizing either form of an SI and a conditional entropy. Since both $F_i$ and $F_T$ are now dynamically learnable, they should be maximized jointly. Only the conditional entropy $\mathcal{H}(F_T|F_i)$ is considered to reflect the causal order of feedforward pass. One way to maximize these terms jointly is the summation, }
\begin{align}\label{eq:add_info}\vspace{-15pt}
    \mathcal{I}^{+}(F_i,F_T) 
    = \mathcal{H}(F_i) + \mathcal{H}(F_T) - \mathcal{H}(F_T|F_i)  
    = \mathcal{H}(F_i) + \mathcal{I}(F_i; F_T) 
\end{align} 
Eq.~\ref{eq:add_info} can be interpreted as maximizing both SI of $F_i$ and $F_T$, while keeping $F_T$ easily identified given $F_i$. The proven neural MI estimator can also be used to estimate and maximize $\mathcal{H}(F_i)$ by rewriting it as $\mathcal{I}(F_i; F_i)$. 
\textcolor{black}{As we are not concerned with the precise value of MI and SI, a more stable Jensen-Shannon MI estimator~\cite{hjelm2018learning} is used, and the loss is by negating the estimation, 
\begin{align}\label{eq::jsd_sp}\vspace{-15pt}
    {\mathcal{L}}^{\text{JSD}}(F_i; F_T) :=
    \mathbb{E}_{p_\text{data}\times\Tilde{p}_{\text{data}}}[\operatorname{sp}\big(T_{\phi}(f_i', f_T)\big)]-
    \mathbb{E}_{p_\text{data}}[-\operatorname{sp}\big(-T_{\phi}(f_i, f_T)\big)]
\end{align}
where $f_i, f_T$ are the features given input from $p_\text{data}$, $f_i'$ is the feature given another input from $\Tilde{p}_\text{data}=p_\text{data}$.  $\operatorname{sp}(\alpha)=\operatorname{log}(1+e^{\alpha})$ is the softplus function.} \medskip\\
\textbf{\textit{Multiplicative Information.}} The SI is a measure of information in the range of $[0,1]$. An alternative is to combine MI and SI and let the SI function as a weighting scheme,
\begin{align} \label{eq:mul_info}
    \mathcal{I}^{\times}(F_i,F_T) 
    = \mathcal{H}(F_i) \times \mathcal{I}(F_i; F_T) 
\end{align} 
Our goal is to jointly maximize each MI and SI. Since both the SI and MI are non-negative, maximizing Eq.~\ref{eq:mul_info} can potentially result in maximizing $\mathcal{H}(F_i)$ and $\mathcal{I}(F_i; F_T)$. \textcolor{black}{Practically, we still use Eq.~\ref{eq::jsd_sp} as the loss of MI and SI. It is always positive, thus minimizing the multiplication of loss can have the same affect as Additive Information that minimizes each term.} \textcolor{black}{Importantly, Multiplicative Information introduces an interesting property where SI functions as a weighting scheme to control MI training process. The MI components of features that have higher SI loss are accordingly weighed more in the penalty function} \textcolor{black}{(more details in Sec.~\ref{sec::ablation})}. In the following sections, we refer to our method as MUSE and use MI$+$SI or MI$\times$SI to specifically refer to the variant of Additive Information or Multiplicative Information.\medskip\\
\textbf{Constraining features with supervision.}
In unsupervised representation learning, the MI estimator is shown to be sensitive to different downstream tasks~\cite{hjelm2018learning,Tschannen2020}. As we tend to construct dependencies between features via the lens of MI, it is essential that all features are aware of the task content. For supervised tasks, we can explicitly provide supervision to guide the learning process and improve the feature quality. To this end, each feature is accompanied by a task-specific objective. We use cross-entropy loss with logits distillation loss for the image classification task and bounding box regression loss along with focal losses on object and class for the object detection task. In Section~\ref{sec::ablation}, we show task-specific loss can improve the overall performance, though merely combining MI with SI in a task-agnostic manner can already significantly outperform other baselines. This indicates the proposed method can effectively introduce a necessary dependency between shallow features and the last feature. \medskip\\
 \textbf{Practical multi-view formulations on MI.} 
In a CNN, deeper layers extract a high-level representation of the data while shallower layers explore local patterns better. 
MUSE follows~\cite{hjelm2018learning} to estimate MI by constructing the views of \textit{global} and \textit{local} structures (see the supplementary material for implementation details).
This formulation can possibly help introduce dependencies among features from different levels, as deeper features are presumably more related to global information and shallower features contain more local information. 

\vspace{-1em}
\section{Experiments}
\vspace{-0.5em}
We conduct extensive experiments on our main target, self-distillation, with various CNN architectures for image classification and object detection. We also include a variety of ablation studies to show the effectiveness of each module. Then we apply MUSE to online and offline distillation and demonstrate its effectiveness.
\vspace{-1em}
\subsection{Image Classification}\label{classification}
\vspace{-0.5em}
We consider a variety of backbone networks -- VGG19~(BN)~\cite{simonyan2015deep}, ResNet~\cite{resnet}, DenseNet~\cite{huang2017densely}, NASNet~\cite{zoph2018learning} and EfficientNet~\cite{pmlr-v97-tan19a} --
on CIFAR100~\cite{Krizhevsky09learningmultiple}, TinyImageNet\footnote{https://tiny-imagenet.herokuapp.com/}, and ImageNet~\cite{imagenet}. For classification, we use labels to calculate cross-entropy at each intermediate features. We also add knowledge distillation loss~\cite{hinton2015distilling} as part of the task-specific loss in this line of experiments.
We will further show in Section~\ref{sec::ablation} that MUSE is necessary for the improvement without knowledge distillation loss and cross-entropy loss.
For CIFAR100 and TinyImageNet, we train the networks with MUSE using SGD with
momentum 0.9, weight decay 5e-4, learning rate initialized as 0.1 and divided by 10 after epoch 75, 130, and 180 for total 200 epochs. \textcolor{black}{For ImageNet, we use SGD with momentum 0.9, weight decay 1e-4, and learning rate initialized as 0.1 and divided by 10 after epoch 30 and 60 for total 90 epochs.} The batch size is 128, 64, and \textcolor{black}{256} for CIFAR100, TinyImageNet, and \textcolor{black}{ImageNet, respectively}. The depth-wise decomposition strategy empirically depends on the architecture, e.g., ResNet has 4 stages, so each stage is a module; VGG is decomposed at the positions of the first 4 maxpooling layers. More details can be found in the supplementary material. \medskip\\
\begin{table*}[t]
\centering
\scalebox{0.5}{
\begin{tabular}{@{}lc@{}c@{}cc@{}c@{}c@{}cc@{}c@{}c@{}cc@{}c@{}c@{}c@{}cc@{}}
\toprule
  \multirow{3}{*}{\textbf{CIFAR100}}
 &\multicolumn{3}{c}{\textbf{Module 1}} & &\multicolumn{3}{c}{\textbf{Module 2}} &
 &\multicolumn{3}{c}{\textbf{Module 3}} &
 &\multicolumn{4}{c}{\textbf{Module 4}} &
 \\
\cmidrule{2-4}\cmidrule{6-8}\cmidrule{10-12}\cmidrule{14-17}
 
 & { BYOT } & { MI+SI }  & { MI $\times$SI }  &
 & { BYOT } & { MI+SI }  & { MI$\times$SI } &
 & { BYOT } & { MI+SI }  & { MI$\times$SI } &
 & { BYOT } & { MI+SI }  & { MI$\times$SI }  & { Baseline }\\
\midrule
VGG19
& { $56.92_{\pm0.15}$ } & { $\bf{60.10_{\pm0.11}}$ } & { ${57.21_{\pm0.14}}$ } &
& { $65.72_{\pm0.19}$ } & { $\bf{68.06_{\pm0.19}}$ } & { ${66.14_{\pm0.17}}$ } &
& { $68.55_{\pm0.17}$ } & { $68.96_{\pm0.18}$ } & { $\bf{69.26_{\pm0.18}}$ } &
& { $69.79_{\pm0.24}$ } & { $71.07_{\pm0.21}$ } & { $ \textcolor{red}{\bf{71.24_{\pm0.18}}}$ } & { $68.57_{\pm0.27}$ } \\
ResNet18
& $67.17_{\pm0.14}$ & $67.14_{\pm0.42}$ & $\bf{68.12_{\pm0.24}}$ &
& $73.27_{\pm0.19}$ & $\bf{74.16_{\pm0.13}}$ & ${73.98_{\pm0.28}}$ &
& $77.14_{\pm0.21}$ & $\bf{77.96_{\pm0.29}}$ & ${77.45_{\pm0.30}}$ &
& $77.86_{\pm0.30}$ & \textcolor{red}{$\bf{78.75_{\pm0.31}}$} & ${78.37_{\pm0.34}}$
& $77.43_{\pm0.36}$ \\
ResNet34
& $65.77_{\pm0.07}$ & $\bf{68.58_{\pm0.37}}$ & ${68.47_{\pm0.31}}$ &
& $75.15_{\pm0.21}$ & $\bf{75.95_{\pm0.23}}$ & ${74.93_{\pm0.33}}$ &
& $78.11_{\pm0.17}$ & $\bf{79.77_{\pm0.20}}$ & ${78.91_{\pm0.21}}$ &
& $78.96_{\pm0.11}$ & \textcolor{red}{$\bf{80.11_{\pm0.19}}$} & ${79.39_{\pm0.16}}$
& $77.56_{\pm0.24}$\\
ResNet50
& $68.86_{\pm0.17}$ & $\bf{71.24_{\pm0.33}}$ & ${69.82_{\pm0.21}}$ &
& $77.71_{\pm0.16}$ & $77.77_{\pm0.27}$ & $\bf{78.04_{\pm0.13}}$ &
& $80.04_{\pm0.12}$ & $\bf{81.25_{\pm0.29}}$ & ${80.26_{\pm0.14}}$ &
& $80.56_{\pm0.16}$ & \textcolor{red}{$\bf{81.44_{\pm0.22}}$} & {${80.63_{\pm0.16}}$} 
& $77.80_{\pm0.23}$\\
ResNet101
& $71.97_{\pm0.23}$ & $72.00_{\pm0.33}$ & $\bf{72.15_{\pm0.21}}$ &
& $75.61_{\pm0.21}$ & $\bf{78.65_{\pm0.34}}$ & ${77.06_{\pm0.14}}$ &
& $78.92_{\pm0.19}$ & $\bf{81.53_{\pm0.23}}$ & ${80.74_{\pm0.32}}$ &
& $79.06_{\pm0.27}$ & \textcolor{red}{$\bf{81.72_{\pm0.37}}$} & ${80.89_{\pm0.17}}$ 
& $77.97_{\pm0.15}$\\
ResNet152 
& $67.72_{\pm0.21}$ & $70.46_{\pm0.33}$ & $\bf{70.92_{\pm0.23}}$ &
& $77.82_{\pm0.24}$ & $\bf{79.53_{\pm0.28}}$ & ${78.38_{\pm0.18}}$ &
& $79.91_{\pm0.17}$ & $\bf{81.83_{\pm0.31}}$ & ${81.77_{\pm0.21}}$ &
& $80.73_{\pm0.19}$ & \textcolor{red}{$\bf{82.09_{\pm0.36}}$} & ${81.69_{\pm0.19}}$ 
& $78.85_{\pm0.26}$\\
NASNet
& ${66.85_{\pm0.21}}$ & $\bf{67.45_{\pm0.33}}$ & $66.50_{\pm0.39}$ &
& $73.73_{\pm0.26}$  & $74.17_{\pm0.19}$ & $\bf{74.43_{\pm0.21}}$ &
& $75.01_{\pm0.31}$  & $\bf{76.81_{\pm0.28}}$ & ${76.41_{\pm0.33}}$ &
& $75.85_{\pm0.29}$  & \textcolor{red}{$\bf{77.11_{\pm0.27}}$} & ${76.87_{\pm0.21}}$ 
& $75.64_{\pm0.41}$\\
EfficientNet-B0
& $70.21_{\pm0.09}$ & $\bf{70.96_{\pm0.16}}$ & $70.83_{\pm0.17}$ &
& $77.28_{\pm0.13}$ & $78.15_{\pm0.16}$ & $\bf{78.40_{\pm0.11}}$ &
& $77.93_{\pm0.08}$ & $78.40_{\pm0.18}$ & $\bf{78.44_{\pm0.12}}$ &
& $78.00_{\pm0.14}$ & $78.56_{\pm0.14}$ & \textcolor{red}{$\bf{78.89_{\pm0.12}}$} 
& $77.61_{\pm0.13}$

\end{tabular}}

\scalebox{0.5}{\begin{tabular}{@{}lc@{}c@{}cc@{}c@{}c@{}cc@{}c@{}c@{}cc@{}c@{}c@{}c@{}cc@{}}
\toprule
 \multirow{3}{*}{\textbf{TinyImageNet}} &\multicolumn{3}{c}{\textbf{Module 1}} & &\multicolumn{3}{c}{\textbf{Module 2}} &
 &\multicolumn{3}{c}{\textbf{Module 3}} &
 &\multicolumn{4}{c}{\textbf{Module 4}} &
  \\
\cmidrule{2-4}\cmidrule{6-8}\cmidrule{10-12}\cmidrule{14-17} 
& { BYOT } & { MI+SI } & { MI$\times$SI }  &
& { BYOT } & { MI+SI } & { MI$\times$SI }  &
& { BYOT } & { MI+SI } & { MI$\times$SI } &
& { BYOT } & { MI+SI } & { MI$\times$SI } & { Baseline }\\
\midrule
ResNet18
& { $42.06_{\pm0.18}$ } & { $42.34_{\pm0.20}$ } & { $\bf{42.54_{\pm0.19}}$ } &
& { $52.05_{\pm0.21}$ } & { $\bf{53.02_{\pm0.23}}$ } & { ${52.49_{\pm0.22}}$ } &
& { $62.31_{\pm0.25}$ } & { $\bf{63.14_{\pm0.19}}$ } & { ${62.98_{\pm0.23}}$ } &
& { $65.60_{\pm0.17}$ } & { $66.72_{\pm0.22}$ } & { \textcolor{red}{$\bf{67.31_{\pm0.20}}$} } 
& { $64.68_{\pm0.30}$ } \\
ResNet34
& $44.02_{\pm0.28}$ & $\bf{45.71_{\pm0.22}}$ & $45.32_{\pm0.24}$ &
& $56.68_{\pm0.31}$ & $\bf{58.39_{\pm0.26}}$ & $58.27_{\pm0.19}$ &
& $66.67_{\pm0.23}$ & $67.48_{\pm0.19}$ & $\bf{67.59_{\pm0.25}}$ &
& $68.44_{\pm0.25}$ & \textcolor{red}{$\bf{69.41_{\pm0.31}}$} & $69.13_{\pm0.22}$
& $66.72_{\pm0.27}$ \\
EfficientNet-B0
& $53.02_{\pm0.08}$ & $53.06_{\pm0.11}$ & $\bf{55.45_{\pm0.14}}$ &
& $62.05_{\pm0.11}$ & $\bf{64.77_{\pm0.16}}$ & $62.51_{\pm0.16}$ &
& $64.91_{\pm0.13}$ & ${65.30_{\pm0.15}}$ & $\bf{66.33_{\pm0.11}}$ &
& $65.50_{\pm0.09}$ & $65.59_{\pm0.10}$ & \textcolor{red}{$\bf{66.41_{\pm0.17}}$}
& $64.55_{\pm0.14}$\\
\end{tabular}}
\scalebox{0.59}{\begin{tabular}{lccccccccccccccccc}
\toprule
\multirow{3}{*}{\textbf{ImageNet}} &\multicolumn{3}{c}{\textbf{Module 1}} & &\multicolumn{3}{c}{\textbf{Module 2}} &
&\multicolumn{3}{c}{\textbf{Module 3}} &
&\multicolumn{4}{c}{\textbf{Module 4}} & \\
\cmidrule{2-4}\cmidrule{6-8}\cmidrule{10-12}\cmidrule{14-17} 
& BYOT & MI+SI & MI$\times$SI  &
& BYOT & MI+SI & MI$\times$SI  &
& BYOT & MI+SI & MI$\times$SI  &
& BYOT & MI+SI & MI$\times$SI  & {Baseline}\\
\midrule
ResNet18
& $41.26$ & $\bf{41.51}$ & $41.24$ &
& $51.94$ & $\bf{52.36}$ & $51.99$ &
& $62.29$ & $63.45$ & $\bf{63.71}$ &
& $69.84$ & $70.35$ & $\textcolor{red}{\bf{70.57}}$
& $69.64$ \\
\bottomrule
\end{tabular}}
\caption{\textbf{Image Classification on CIFAR100 (top), TinyImageNet (middle) and ImageNet (bottom).} Top-1 accuracy averaged over 3 runs, higher is better. The best of each Module is in \textbf{bold}. The best for an architecture is in \textcolor{red}{red}. MI+SI: Additive Information, MI$\times$SI: Multiplicative Information. MUSE outperforms on all modules and the baseline.}
\vspace{-1em}
\label{table:compare_byot}
\end{table*}

\vspace{-2em}
\paragraph{\textbf{Comparison with BYOT.} }
\vspace{-0.5em}

MUSE is related to BYOT~\cite{Zhang_2019_ICCV} in the form of module decomposition. Therefore, we decompose the backbone into 4 modules to fairly compare with BYOT~\cite{Zhang_2019_ICCV}.
 Table~\ref{table:compare_byot} shows that MUSE consistently outperforms BYOT significantly for all modules on all networks.
 We posit that MUSE introduces the feature dependencies and enhances feature expressivity by maximizing MI and SI. Meanwhile BYOT minimizes the L2 loss in the feature space which leads to an undesirable outcome that forces the feature distributions to be identical (discussed in Section~\ref{sec::MUSE}). It poses an unexpected regularization and worsens the performance. Further empirical results and discussion on MUSE as a functional proxy to introduce feature dependency can be found in Table~\ref{table:discrepancy} and Section~\ref{sec::ablation}. Interestingly, for the two variants of MUSE, MI$\times$SI is shown to be more likely to perform better on VGG19 and EfficientNet-B0, while MI$+$SI tends to outperform on other ResNets and NASNet. For those networks where MI$\times $SI performs worse, the performance gaps between layers are shown to be larger, and therefore different features are not comparably informative. A direct multiplication may allow the network to ignore some shallower features. 
  \vspace{-1em}
\paragraph{\textbf{Comparison with self-distillation.}}
\vspace{-0.5em}
 We compare MUSE to other state-of-the-art SD approaches: Class-wise Self-knowledge Distillation (CS-KD)~\cite{Yun_2020_CVPR}, On-the-fly Native Ensemble (ONE)~\cite{lan2018knowledge}, Data-Distortion Guided Self-Distillation (DDGSD)~\cite{ddsgd2019}, and Feature Refinement via Self-Knowledge Distillation (FRSKD)~\cite{ji2021refine}. We report the accuracy of the last module for MUSE and BYOT. Table~\ref{table:comparison_sd} indicates that MUSE can beat all other SD methods. Note that MUSE serves as the feature discrepancy function and is therefore orthogonal to other SD methods. It can be applied to these methods and potentially further improve their performance.

\begin{table}[t]
\centering
\caption{\textbf{Self-distillation comparison on CIFAR100.} Evaluated by top-1 accuracy over 3 runs, higher is better. Best result is in bold. $\dagger$: reported result from \cite{ji2021refine}.}
\scalebox{0.7}{\begin{tabular}{lccccccccccc}
\toprule
 \textbf{Method}
&{\textbf{Baseline}}
&{\textbf{CS-KD}}
&{\textbf{ONE}}
&{\textbf{DDGSD}}
&{\textbf{BYOT}}
&{\textbf{FRSKD}\textsuperscript{$\dagger$}}
&{\textbf{MI+SI (ours)}}
&{\textbf{MI$\times$SI (ours)}}\\
\midrule
ResNet18 & $77.43_{\pm0.36}$ & $78.01_{\pm0.11}$ & $77.03_{\pm0.21}$ & $77.88_{\pm0.32}$  & $77.86_{\pm0.30}$ & $77.71_{\pm0.14}$& $\bf{78.75_{\pm0.31}}$ & $78.37_{\pm0.34}$ \\
DenseNet121  & $77.01_{\pm0.06}$& $78.25_{\pm0.12}$ & $76.88_{\pm0.29}$ & $78.04_{\pm0.19}$  & $78.15_{\pm0.14}$ &$-$& ${78.45_{\pm0.07}}$ & $\textbf{78.56}_{\pm0.11}$ \\
\bottomrule
\end{tabular}}
\vspace{-1em}
\label{table:comparison_sd}
\end{table}

\paragraph{Model Compression.}

\vspace{-1em}
\begin{wraptable}[11]{r}{5.8cm}
\centering
\vspace{-10pt}
\setlength{\tabcolsep}{2pt}
\caption{\textbf{Compact networks by MUSE.}}
\label{table:compression}
\scalebox{0.55}{
\begin{tabular}{lcccccccc}
\toprule
  \multirow{3}{*}{\textbf{CIFAR100}}&&\multicolumn{3}{c}{\textbf{Baseline}}&  &\multicolumn{3}{c}{\textbf{MUSE}} \\
\cmidrule{3-5}\cmidrule{7-9}
  &
 & top-1 & params & FLOPs & 
 & top-1 & params & FLOPs \\
\midrule
VGG19 &
& $68.57$ & $20.0$M & $20499$M &
& $69.26$ & $2.3$M & $2380$M\\
ResNet18 &
& $77.43$ & $11.2$M & $256$M &
& $77.96$ & $2.8$M & $248$M\\
ResNet34 &
& $77.56$ & $21.3$M & $393$M &
& $79.77$ & $8.2$M & $380$M\\
ResNet50 &
& $77.80$ & $23.4$M & $431$M &
& $78.04$ & $1.4$M & $370$M\\
ResNet101 &
& $77.97$ & $42.4$M & $507$M &
& $78.65$ & $1.4$M & $370$M\\
ResNet152 &
& $78.85$ & $58.0$M & $636$M &
& $79.53$ & $2.5$M & $441$M\\
NASNet &
& $75.64$ & $5.1$M & $241$M &
& $76.81$ & $1.9$M & $238$M\\
EfficientNet-B0 &
& $77.61$ & $4.0$M & $454$M &
& $78.40$ & $0.8$M & $290$M\\
\midrule
\multirow{3}{*}{\textbf{TinyImageNet}}&&\multicolumn{3}{c}{\textbf{Baseline}}&  &\multicolumn{3}{c}{\textbf{MUSE}}\\
\cmidrule{3-5}\cmidrule{7-9}
 &
 & top-1 & params & FLOPs & 
 & top-1 & params & FLOPs \\
\midrule
 ResNet34 &
& $66.72$ & $21.3$M & $1543$M &
& $67.59$ & $8.2$M & $1520$M\\
 EfficientNet-B0 &
& $64.55$ & $4.0$M & $454$M &
& $64.77$ & $0.8$M & $290$M\\

\bottomrule
\end{tabular}}
\vspace{-1em}
\end{wraptable}

In addition to improving accuracy as discussed in Section~\ref{classification}, MUSE can intrinsically achieve model compression without sacrificing performance. Since intermediate features are extracted for predictions, the models can be compressed if the prediction performance from the intermediate features outperforms the baseline. For example, Module 3 from VGG19 has higher top-1 accuracy compared to the baseline on CIFAR100 (69.26\% vs. 68.57\%). Hence, by training the model with MUSE, modules after Module 3 can be discarded in inference while achieving higher accuracy with less parameters and computation.
Table~\ref{table:compression} shows the comparisons between the compressed models and the corresponding baseline models shown in Table~\ref{table:compare_byot}. All the compressed models achieve higher top-1 accuracy while the number of parameters and floating-point operations per second (FLOPs) are significantly reduced (up to $30.3\times$ and $8.6\times$, respectively). The compression ratio can be traded off for accuracy by changing the layers where intermediate features are extracted in MUSE. This approach is orthogonal to other model compression methods such as pruning and quantization and can be combined with them to further compress the models.\\

\vspace{-1em}
\subsection{Object Detection}

\begin{wraptable}[5]{r}{5.5cm}
\centering
\vspace{-12pt}
\setlength{\tabcolsep}{2pt}
\caption{\textbf{Object Detection}, mAP with 0.5 IoU. Higher is better.}
\label{table:detection}
\scalebox{0.6}{\begin{tabular}{ccccccccc}
\toprule
 \textbf{Networks} 
 & YOLOv5-S 
 & YOLOv5-M 
 & YOLOv5-L 
 & YOLOv5-X \\
\midrule
Baseline
& $0.549_{\pm0.006}$ &  $0.619_{\pm0.001}$ &  $0.641_{\pm0.001}$ &  $0.663_{\pm0.002}$ \\
\midrule
MI+SI
& $0.543_{\pm0.001}$ &  $0.627_{\pm0.002}$ &  $0.655_{\pm0.003}$ & $0.666_{\pm0.001}$ \\
\midrule
MI$\times$SI
& $\bf{0.559_{\pm0.001}}$ &  $\bf{0.629_{\pm0.002}}$ & $\bf{0.659}_{\pm0.003}$& $\bf{0.672}_{\pm0.001}$ \\

\bottomrule
\end{tabular}}
\vspace{-1em}
\end{wraptable}

We apply MUSE on the Yolov5 family for object detection on COCO~\cite{lin2014microsoft} dataset. Yolov5 models consist of a backbone and a head. The backbone is decomposed into two modules: the last 3 components (Conv, SPP, C3) and the other components in front of them. We estimate MI and SI using the features out of the two modules similar to Section~\ref{classification}. A bottleneck layer is concatenated after the first backbone module followed by a detection head (same as the original Yolov5). 
All networks are trained from scratch using SGD with momentum $0.937$, weight decay $0.01$, and the learning rate initialized as 0.01 and decayed to 0.002 in a sinusoidal ramp. We follow the batch size as official Yolov5 repo\footnote{https://github.com/ultralytics/yolov5}.  
As shown in Table~\ref{table:detection}, MI$\times$SI variant can improve the detection performance of all Yolov5 models. The MI$+$SI variant, on the other hand, is able to improve the mAP of Yolov5-M and Yolov5-L. 
Compared to image classification, low-level features like spatial locality is critical for object detection. We hypothesize that the weighting scheme in MI$\times$SI possibly attaches more importance to those shallow features, rather than equivalently optimizing all terms as in MI$+$SI.

\begin{figure}[t]
\begin{center}

\includegraphics[width=1\textwidth]{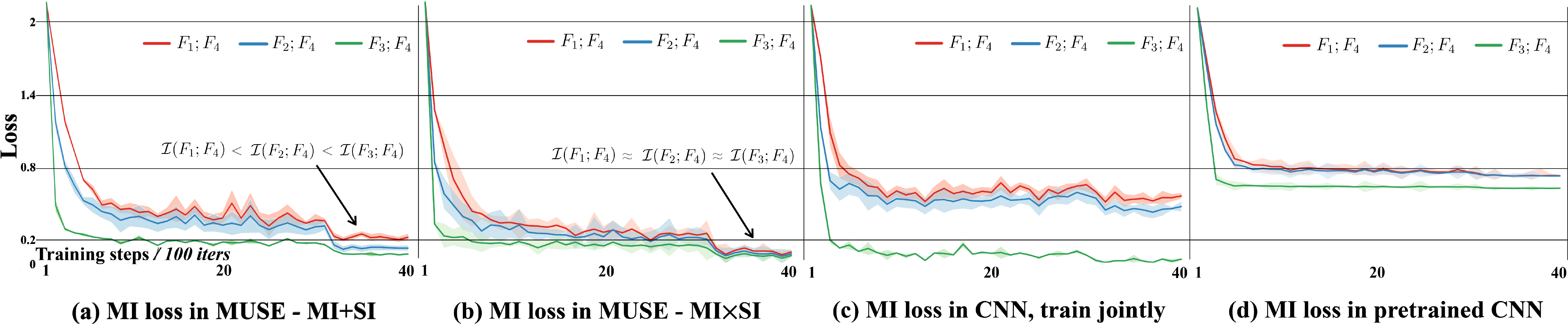}

\end{center}
\vspace{-1em}
\caption{\textbf{Training curves of MI loss on CIFAR100 EfficientNet-B0.} X-axis: training steps (per 100 iterations), y-axis: loss (always positive, lower means higher MI). \textcolor{layer1}{Red}, \textcolor{layer2}{blue}, \textcolor{layer3}{green} denote MI loss between $F_4$ and \textcolor{layer1}{$F_1$}, \textcolor{layer2}{$F_2$}, \textcolor{layer3}{$F_3$}. The value of each MI loss is comparable as its implementation is same for all. }
\label{fig::MIs_all}\vspace{-1em}
\end{figure}

 \begin{wrapfigure}[16]{r}{4.3cm}
 \vspace{-33pt}
 
 \begin{center}

   \includegraphics[width=\textwidth]{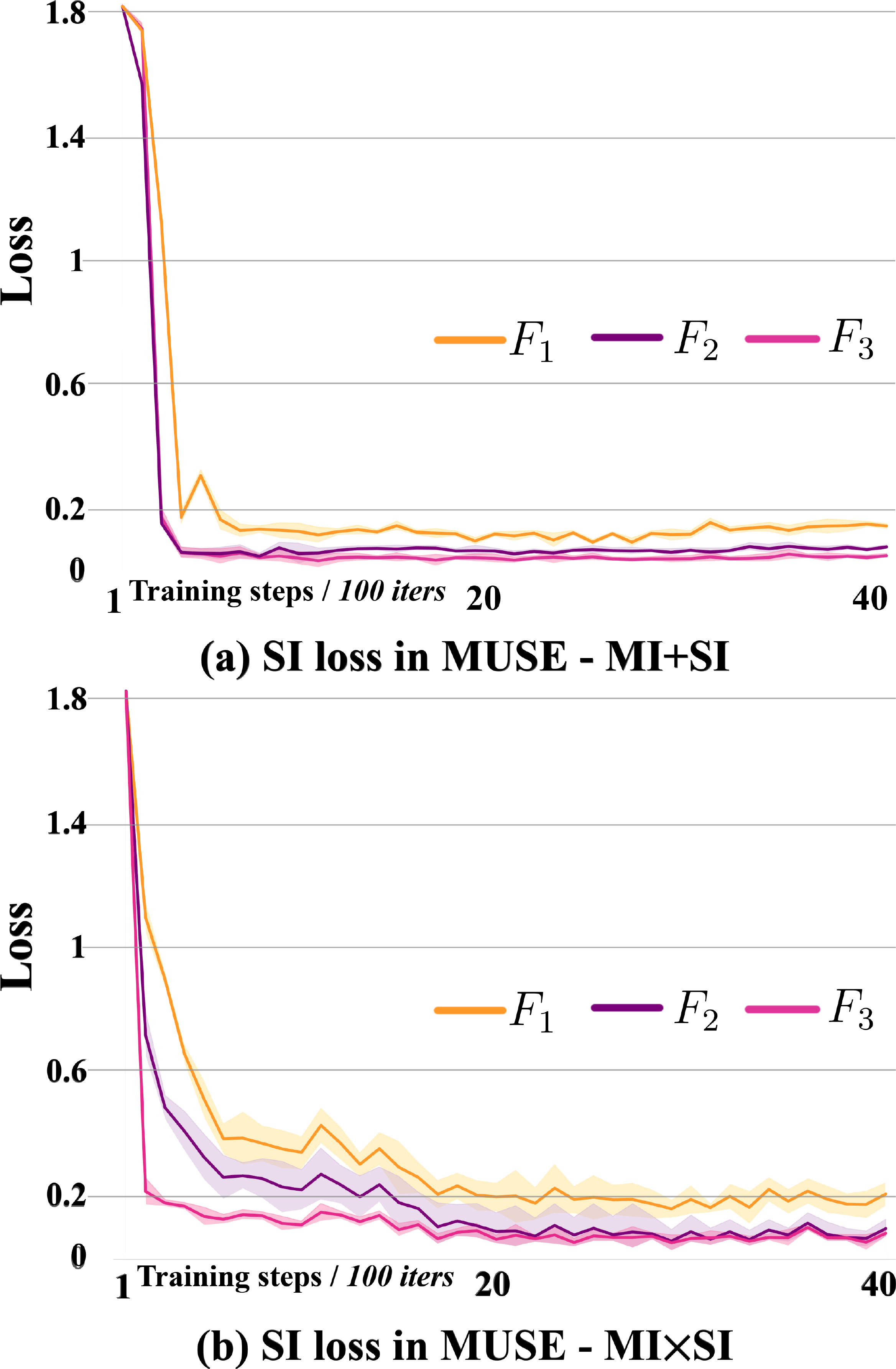}

 \end{center}
 \vspace{-1em}
 \caption{\textbf{SI loss on CIFAR100 EfficientNet-B0} of \textcolor{s1}{$F_1$}, \textcolor{s2}{$F_2$} and \textcolor{s3}{$F_3$}.}
 \label{fig::SIs_all}\vspace{-1em}
\end{wrapfigure}

\vspace{-1em}
\subsection{Ablation study} \label{sec::ablation}

\vspace{-3em}
\textcolor{black}{
\paragraph{MI \& SI interaction in MUSE.}
Though the MI and SI in our implementation are not precisely estimated, the decrease of the loss can indicate the relative value of MI and SI. 
In Fig.~\ref{fig::MIs_all}\textcolor{red}{(a)} and \ref{fig::MIs_all}\textcolor{red}{(b)}, all MI losses in MI$\times$SI converge to a similar level, whereas converged MI losses in MI$+$SI are diverse (significantly higher for shallower features). We can see in Fig.~\ref{fig::SIs_all} that shallower features consistently have higher SI losses (lower SI) for both MI$+$SI and MI$\times$SI.  Therefore for MI$\times$SI where MI loss is weighted by SI loss (always positive), the optimization of shallow features is boosted . 
It also explains why MI$\times$SI shows superior performance in object detection (Table~\ref{table:detection}), as the information of lower-level feature is better captured in MI$\times$SI where shallow features are weighted with more importance. Hence, shallow features tend to have lower MI losses (higher MI) in MI$\times$SI than MI$+$SI after convergence. Dense prediction like object detection may prefer such well-trained multi-scale feature information.
If MI is the regularizer for CNN training, we can further interpret SI as the regularizer for the MI training. 
}

\vspace{-1.5em}
\textcolor{black}{
\paragraph{MI in MUSE and conventional CNN.}
A typical strategy (including MUSE) is to estimate and maximize MI jointly with CNN training. To compare with MI in baseline CNNs, we evaluated on MI with trainable CNN (Fig.~\ref{fig::MIs_all} \textcolor{red}{(c)}) and pretrained CNN (Fig.~\ref{fig::MIs_all} \textcolor{red}{(d)}) for a full picture of MI in baseline CNNs (EfficientNet in this example). Note that the implementation of Fig.~\ref{fig::MIs_all} \textcolor{red}{(c)} is the same as only MI in Table~\ref{table:discrepancy}. The last MI loss between $F_3;F_4$ can still converge to a very low value but the other two are poorly optimized, indicating the effectiveness of SI in MUSE to control the MI training process. In Fig.~\ref{fig::MIs_all} \textcolor{red}{(d)}, the MI losses in the pretrained CNN cannot converge to a stage as low as the other three. We speculate it is because---(1) the MI between intermediate features of conventional CNNs are possibly much lower; (2) The MI estimators cannot work properly without trainable parameters of the backbone.}

\begin{wraptable}[13]{r}{5cm}
\centering
\vspace{-18pt}
\setlength{\tabcolsep}{2pt}
\caption{\textbf{Different terms in MUSE.}}
\label{table:discrepancy}
\scalebox{0.62}{\begin{tabular}{lcccccccccc}
\toprule
\textbf{Backbone} & {\textbf{ResNet18}} & {\textbf{EfficientNet-B0}}  \\
\midrule
Baseline
& $77.43_{\pm0.36}$ 
& $77.61_{\pm0.13}$ & \\
\midrule
CE
& $77.66_{\pm0.27}$ 
& $77.94_{\pm0.11}$ & \\
CE + KD
& $77.98_{\pm0.21}$ 
& $78.06_{\pm0.18}$ & \\
\midrule
L2
& $77.18_{\pm0.23}$ 
& $77.43_{\pm0.18}$ & \\
L2 + CE
& $77.65_{\pm0.25}$ 
& $77.51_{\pm0.16}$ & \\
L2 + CE + KD
& $77.86_{\pm0.30}$ 
& $78.00_{\pm0.14}$ & \\
\midrule
MI
& $77.95_{\pm0.31}$ 
& $77.67_{\pm0.26}$ & \\
\color{black}SI
& \color{black}$77.81_{\pm0.44}$ 
& \color{black}$77.69_{\pm0.25}$ & \\
\color{black}MI + CE
& \color{black}$78.07_{\pm0.33}$ 
& \color{black}$78.07_{\pm0.11}$ & \\
\color{black}MI + CE + KD
& \color{black}$78.22_{\pm0.21}$ 
& \color{black}$78.35_{\pm0.13}$ & \\
(MI+SI)
& $78.14_{\pm0.25}$ 
& $77.83_{\pm0.19}$ & \\
(MI$\times$SI)
& $78.06_{\pm0.29}$ 
& $77.91_{\pm0.24}$ & \\
(MI+SI) + CE
& $78.63_{\pm0.26}$ 
& $78.46_{\pm0.17}$ & \\
(MI$\times$SI) + CE
& $78.35_{\pm0.24}$ 
& $78.81_{\pm0.12}$ & \\
(MI+SI) + CE + KD
& $\bf{78.75_{\pm0.31}}$ 
& $78.56_{\pm0.14}$ & \\
(MI$\times$SI) + CE + KD
& $78.37_{\pm0.34}$ 
& $\bf{78.89_{\pm0.12}}$ & \\
\bottomrule
\end{tabular}}
\vspace{-1em}
\end{wraptable}

\vspace{-1em}
\paragraph{Effectiveness of different modules.}
In Section~\ref{classification}, we empirically show MUSE outperforms other SOTA SD methods. As our objective incorporates multiple terms including classification loss (cross-entropy) and knowledge distillation loss (KL divergence between logits), we conduct experiments with individual terms as in Table~\ref{table:discrepancy}. We report the top-1 accuracy at Module 4 with 3 runs. CE and KD denote cross-entropy and knowledge distillation loss between intermediate features. We decompose the networks into 4 modules and all terms (CE, KD, L2, MI, and MUSE) are calculated between Module 1-3 and Module 4.  
In Table~\ref{table:discrepancy}, we can observe: \textbf{(1)} CE and KD can improve the overall performance without other feature discrepancy functions; \textbf{(2)} L2 loss itself hurts the original performance, while MI, MI+SI, and MI$\times$SI can further improve the performance with CE and KD. It corroborates with the argument in Section~\ref{sec::MUSE} that MI serves as a more functional proxy to introduce dependencies between intermediate features; \textbf{(3)} MUSE, including both MI+SI and MI$\times$SI, outperforms MI, indicating the effectiveness of combining MI and SI; \textbf{(4)} MUSE can improve the network even without CE and KD, showing that MI and SI can learn more useful features without explicit supervision to the intermediate features; \textbf{(5)} Though MUSE solely improves the performance, adding CE \& KD together provides the maximum improvement. MI estimators are sensitive to the tasks, therefore explicit supervision likely helps introduce meaningful feature dependencies for specific tasks; \textcolor{black}{\textbf{(6)} Using MI or SI solely provides marginal improvement, while combining them performs the best.}
\vspace{-1em}
\paragraph{Advantages over other discrepancy functions. }
 \begin{wraptable}[8]{r}{4.3cm}
 \vspace{-0.5em}
 \setlength{\tabcolsep}{2pt}
 \vspace{-7pt}\caption{\textbf{Feature discrepancy.}}\label{tab:diff_dis}
 \scalebox{0.7}{
\begin{tabular}{lcccccc}
\toprule
\textbf{Backbone} & {\textbf{ResNet18}} & {\textbf{EfficientNet-B0}} \\
\midrule
Baseline
& $77.43_{\pm0.36}$ 
& $77.61_{\pm0.13}$ & \\
L2
& $77.65_{\pm0.25}$ 
& $77.51_{\pm0.16}$ & \\
MMD
& $76.22_{\pm0.19}$  
& $77.03_{\pm0.16}$ & \\
adversarial
& $78.18_{\pm0.18}$ 
& $78.25_{\pm0.14}$ & \\
VID
& $76.97_{\pm0.54}$ 
& $77.37_{\pm0.35}$ &\\
MI+SI
& $\bf{78.63_{\pm0.26}}$ 
& $78.46_{\pm0.17}$ & \\
MI$\times$SI
& $78.35_{\pm0.24}$ 
& $\bf{78.81_{\pm0.12}}$ & \\
\bottomrule
\end{tabular}}
\vspace{-1em}
  \end{wraptable}
Prior distillation works introduce feature dependencies via different feature discrepancy functions, such as L2 loss \cite{Romero15fitnets:hints,  DBLP:conf/iclr/ZagoruykoK17}, Maximum Mean Discrepancy (MMD)~\cite{NST2017}, adversarial loss~\cite{chung2020feature} and VID (a type of MI estimated by variational bound~\cite{ahn2019variational}). To demonstrate the effectiveness of MUSE on SD, we apply these feature discrepancy functions to introduce the feature dependencies and compare to our proposed MUSE in Table~\ref{tab:diff_dis}. All networks incorporate CE loss to provide explicit supervision. KD loss is not included in Table~\ref{tab:diff_dis} to better show the improvement from the feature discrepancy functions. We observe a consistent improvement of MUSE over other functions: \textbf{(1)} L2 loss, often used in offline distillation, cannot consistently improve performance. It demonstrates our argument that enforcing the features to be identical (as L2 loss does) cannot introduce useful dependencies in SD; \textbf{(2)} MMD is the same as L2 that its minimum is obtained if and only if the two features are identical; \textbf{(3)}
Adversarial loss differently shows significant improvement over baselines.  Minimizing L2 loss is equivalent to minimizing the KL divergence between two features $D_\text{KL}(p_1||p_2)$ ($p$ is the probability density), while minimizing adversarial loss is equivalent to minimizing the Jensen–Shannon (JS) divergence~\cite{NIPS2014_5423} between two features. It is a symmetric version of KL divergence that $D_\text{JS}(p_1||p_2) = D_\text{JS}(p_2||p_1)$. As both features are not known, a symmetric discrepancy function without assumption on the direction is possibly preferred. This also explains its empirical success in online distillation~\cite{chung2020feature}. Yet, its minimum is obtained if and only if two features are identical, which 
weakens its faculty in SD; \textbf{(4)} MUSE outperforms MI estimated by variational bound~\cite{ahn2019variational}, as this bound of MI only holds in the offline distillation where the pretrained teacher is static (refer to Eq.3 and 4 in VID~\cite{ahn2019variational}). \medskip\\
\textbf{Extension to Offline / Online Distillation.}
We have shown that MUSE necessarily improves the SD framework on image classification and object detection. We further investigate its potential application on offline and online distillation where the features are from different CNNs. MUSE can be readily applied in this scenario by replacing the own last feature of a CNN with the last feature of another teacher network. We also follow our previous experimental setting to decompose the student network into four modules. To establish a fair comparison, we do not include CE and KD loss for intermediate features, but only calculate MUSE between intermediate features of the student network and the last feature of the teacher network. We follow a traditional strategy to add KD loss on the last layer of the student network. For online distillation, we consider two settings: two identical networks and two different networks. We report the average accuracy for two identical networks. For offline distillation, we use the fixed last feature of the teacher network to calculate MUSE. In Table~\ref{tab::off_on_line}, we can observe consistent improvement from MUSE for online distillation, and comparable performance with the state-of-the-art for offline distillation. 
\begin{table}
\setlength{\tabcolsep}{2pt}
\begin{tabular}{ cc } 
\resizebox{0.58\textwidth}{!}{
\begin{tabular}{lcccccccccccc}
\toprule
 {\textbf{Net1 /}}
&{\textbf{resnet20 /}} 
&{\textbf{resnet56 /}}
&{\textbf{resnet20 /}} 
&{\textbf{ShuffleNetV1}~\cite{8578814}\textbf{ /}} \\
{\textbf{Net2}}
&{\textbf{resnet20 }} 
&{\textbf{resnet56 }}
&{\textbf{resnet56}} 
&{\textbf{WRN-40-2}~\cite{Zagoruyko2016WRN}} \\
\midrule
Baseline
& $69.06$ 
& $72.34$ 
& $69.06 / 72.34$ 
& $70.50 / 75.61$ \\
KD
& $70.51$  
& $75.24$  
& $70.11/74.69$  
& $70.50 / 75.61$ \\
DML~\cite{zhang2017deep}
& $70.84$  
& $75.63$  
& $71.13/74.97$  
& $75.89/78.16$ \\
KDCL~\cite{Guo_2020_CVPR}
& $70.23$  
& $75.28$ 
& $70.36/74.83$ 
& $74.79 / 77.53$ \\
AFD~\cite{chung2020feature}
& $70.63$  
& $75.40$  
& $71.22/75.12$  
& $75.39/77.13$  \\
MI+SI
& $\bf{71.02_{\pm0.14}}$  
& $\bf{76.02_{\pm0.13}}$ 
& $\bf{71.34_{\pm0.16}}/\bf{75.55_{\pm0.08}}$ 
& $76.65_{\pm0.19}/\bf{78.51_{\pm0.1 7}}$ \\
MI$\times$SI
& $70.71_{\pm0.16}$  
& $75.80_{\pm0.15}$  
& $71.05_{\pm0.11}/75.17_{\pm0.10}$  
& $\bf{76.80_{\pm0.21}}$$/78.34_{\pm0.15}$ \\
\bottomrule
\end{tabular}
}
&
\resizebox{0.36\textwidth}{!}{
\begin{tabular}{lcccccccccc}
\toprule
 \textbf{Teacher} 
&{\textbf{resnet56 (72.34)}} 
&{\textbf{resnet110 (74.31)}}\\
 \textbf{Student} 
&{\textbf{resnet20}} 
&{\textbf{resnet32 }}\\
\midrule
Baseline
& $69.06$ 
& $71.14$ \\
KD~\cite{hinton2015distilling}
& $70.66$ 
& $73.08$ \\
FitNets~\cite{Romero15fitnets:hints}
& $69.21$ 
& $71.06$  \\
VID~\cite{ahn2019variational}
& $70.38$ 
& $72.61$ \\ 
CRD~\cite{Tian2020Contrastive}
& $71.16$ 
& $\bf{73.48}$\\ 
MI+SI
& $\bf{71.30_{\pm0.09}}$ 
& $73.34_{\pm0.17}$ \\
MI$\times$SI
& $71.27_{\pm0.22}$ 
& $\bf{73.48_{\pm0.29}}$\\ 
\bottomrule
\end{tabular}
} \\
\footnotesize{\textbf{(a) Online Distillation}.} & \footnotesize{\textbf{(b) Offline Distillation}.} \\
\end{tabular}
\vspace{-0.5em}
\caption{\textbf{Traditional distillation with MUSE.} Evaluated by top-1 accuracy. (a) Two settings are considered in online distillation---two identical networks (averaged accuracy is reported) and two different networks; (b) Pretrained teacher networks in offline distillation are static.}
\label{tab::off_on_line}
\vspace{-1em}
\end{table}

\vspace{-1em}

\section{Conclusion}
\vspace{-1em}
We propose a novel feature discrepancy function---MUSE, based on effective neural estimators of MI and SI.
We present two variants of MUSE to combine MI and SI. We argue and empirically demonstrate on extensive experiments that MUSE is a more effective feature discrepancy function for knowledge distillation. Especially on self-distillation, MUSE necessarily introduces dependencies among features in a CNN, thereby significantly improving the performance and obtaining more compact yet comparably performant subnetworks. MUSE shows superior performance on image classification and object detection. 
By drawing the features from different levels, MUSE can possibly be extended to architectures like RNN or attention models. The \textit{level} may not necessarily be the depth of the network, rather time steps or sequential order. We leave these as future work on improving MUSE.



\bibliography{egbib}
\end{document}